\pdfoutput=1

\documentclass[11pt]{article}

\usepackage[]{acl}

\usepackage{times}
\usepackage{latexsym}
\usepackage{booktabs}
\usepackage{arydshln}
\usepackage{multirow}
\usepackage[normalem]{ulem}

\usepackage[T1]{fontenc}
 
\usepackage[utf8]{inputenc}

\usepackage{microtype}

\usepackage{inconsolata}
\usepackage{graphicx} 
\usepackage{graphics}
\usepackage{amsmath}

\DeclareMathOperator*{\argmin}{arg\,min}

\title{Revisiting Demonstration Selection Strategies in In-Context Learning}

\author{%
  Keqin Peng$^{1}$,
  Liang Ding$^{2}$\thanks{~~Corresponding Authors.},
  Yancheng Yuan$^{3*}$\\
  \textbf{Xuebo Liu}$^{4}$,
  \textbf{Min Zhang}$^{4}$,
  \textbf{Yuanxin Ouyang}$^{1}$,
  \textbf{Dacheng Tao}$^{5}$\\
  $^{1}$Beihang University $^{2}$The University of Sydney 
  $^{3}$The Hong Kong Polytechnic University\\ 
  $^{4}$Harbin Institute of Technology, Shenzhen $^{5}$Nanyang Technological University\\
  \texttt{keqin.peng@buaa.edu.cn},
  \texttt{liangding@gmail.com}}

\begin{document}
\maketitle
\begin{abstract}
Large language models (LLMs) have shown an impressive ability to perform a wide range of tasks using in-context learning (ICL), where a few examples are used to describe a task to the model. However, the performance of ICL varies significantly with the choice of demonstrations, and previous research usually focuses on the data aspect ignoring the model's effect. In this work, we first revisit the factors contributing to this variance from \textcolor{black}{the model aspect, and find that the demonstration choice is both data- and model-dependent. We further propose a conjecture that \textit{the performance of a demonstration positively correlates with its contribution to the model's understanding of the test samples}, and accordingly propose a data- and model-dependent demonstration selection method, \textbf{TopK + ConE}.}
Empirically, our method yields consistent improvements in both language understanding and generation tasks with different model scales. Further analyses confirm that, besides the generality and stability under different circumstances, our method provides a unified explanation for the effectiveness of previous methods. Code is publicly available at \url{https://github.com/Romainpkq/revisit_demon_selection_in_ICL}.
\end{abstract}

\section{Introduction}
Large language models (LLMs,~\citealp[]{ouyang2022training, touvron2023llama}) have achieved widespread success across many NLP tasks~\cite{zhong2023chat,Peng2023ChatGPT4MT,Lu2023EAPrompt} due to their remarkable emergent abilities~\cite{DBLP:journals/tmlr/WeiTBRZBYBZMCHVLDF22}. One of the most exciting emergent abilities is in-context learning (ICL,~\citealp[]{DBLP:conf/nips/BrownMRSKDNSSAA20}), which utilizes only a few input-output examples to help LLMs make better predictions~\cite{dong2022survey}. ICL has shown its effectiveness in eliciting LLMs' advanced capabilities and has (almost) become a common practice in tackling complex tasks. However, prior work~\cite{liu2022makes, lu-etal-2022-fantastically} has found that ICL is very sensitive to the choice of in-context examples and their order in the prompt, and even small changes can result in large variance~\cite{iter-etal-2023-context}.

The sensitivity of ICL motivates researchers to explore methods to identify stable and high-performing demonstrations. 
Influenced by the success of leveraging a retrieval module to augment neural networks~\cite{hashimoto2018retrieve}, the retrieval module has become a standard module in the ICL framework for retrieval demonstrations from a dataset~\cite{liu2022makes,rubin-etal-2022-learning}. Extensive research has been conducted to search for demonstrations similar to the test samples~\cite{liu2022makes, Selective_Annotation, robertson2009probabilistic}. For example, \citet{liu2022makes} proposed to select the samples that are closer to the test sample in the embedding space as in-context examples, and ~\citet{ robertson2009probabilistic} found that choosing the high word-overlap samples can also improve the ICL performance. 

\begin{figure}[t!]
    \centering
    \includegraphics[width=0.9\linewidth]{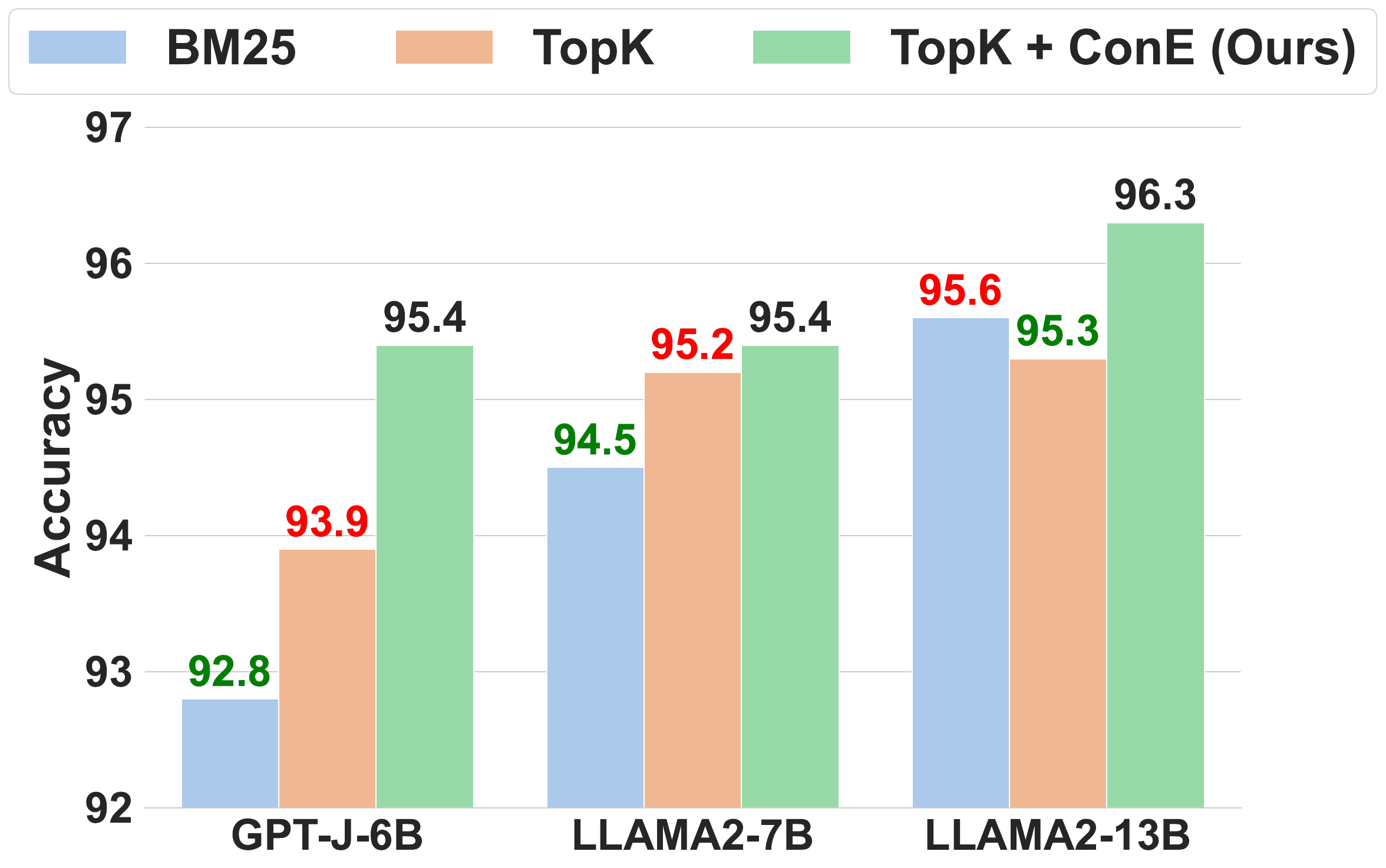}
    \caption{\textbf{The different 8-shot performance of data-dependent methods (BM25 and TopK) and Our methods} in SST-2. The colour in the number represents the relative performance between BM25 and TopK. We see that: 1) The data-dependent methods can not obtain optimal demonstrations under different models; 2) Our data- and model-dependent methods can achieve consistent improvement across different models.}
    \label{fig:pre}
\end{figure}

Despite empirical success to some extent, the above methods usually only focus on the test data, overlooking the impact of models.
To figure out what factors influence the choice of demonstrations, we revisit the performance of ICL from 
the model aspect, \textcolor{black}{and accordingly propose a conjecture to understand the effective demonstrations}. 
Specifically, we investigate ICL performance across different retrieval modules and inference models in \S\ref{subsec:influence_factors}.
Experimental results show that the ICL performance can largely vary with different models even with the same demonstrations (see Figure~\ref{fig:pre} as an example), indicating that \textbf{\textit{the choice of demonstration is not only dependent on test data but also on the retrieval modules and inference models}}. 
\textcolor{black}{We further propose a corresponding conjecture that \textit{effective demonstrations are those that enhance the inference model's understanding of the test input}, and the comparison results between shuffled test input and original test input demonstrate that \textbf{the ICL performance positively correlates with model's understanding of the test samples}.}

Based on the above conjectures, we accordingly propose a demonstration selection method, denoted as~\textbf{TopK+ConE}. 
Specifically, we initially employed the TopK~\cite{liu2022makes} method to narrow down the pool of demonstration candidates, followed by ranking these candidates based on the \textbf{con}ditional \textbf{e}ntropy (estimated by the model itself) of the test sample input. 
Extensive experiments demonstrate the effectiveness of our method across different model scales. Further analyses show the universality and robustness, and provide a unified view of why previous demonstration selection methods work.
Our \textbf{contributions} are summarized as follows:
\begin{itemize}
  \item To the best of our knowledge, we are the first to study the impact of models on the demonstration selection methods. We substantiate that the choice of demonstrations is not only dependent on the test data but also on the retrieval module and inference model.
  \item We build the connection between ICL performance with the model's understanding of test inputs. Our findings reveal that ICL performance positively correlates with the model’s understanding of the test samples.
  \item We propose a data- and model-dependent method \textbf{TopK+ConE} to effectively enhance the models' understanding of test input via reducing the conditional entropy of test input under the inference model.
  \item We achieve state-of-the-art performance on a series of tasks, and prove the effectiveness and universality of our method. Hopefully, our proposed best practice can be employed by more LLM participants.
\end{itemize}

\section{Revisiting Demonstrations Selection}
While in-context learning (ICL,~\citealp[]{brown2020language,dong2022survey}) has shown its impressive few-shot performance, recent work has found that LLMs are very sensitive to the selected examples leading to large variances in performance~\cite{zhao2021calibrate}. Although many advanced ICL strategies~\cite{robertson2009probabilistic,liu2022makes,wu-etal-2023-self} have been proposed to select effective demonstrations, \textcolor{black}{why these demonstrations work and what factors influence their selection have not been fully studied. In this section, we first explore the influencing factors to the demonstration selection and correspondingly propose a conjecture to understand the effective demonstrations.}
\begin{figure}[t!]
    \centering
    \includegraphics[width=0.8\linewidth]{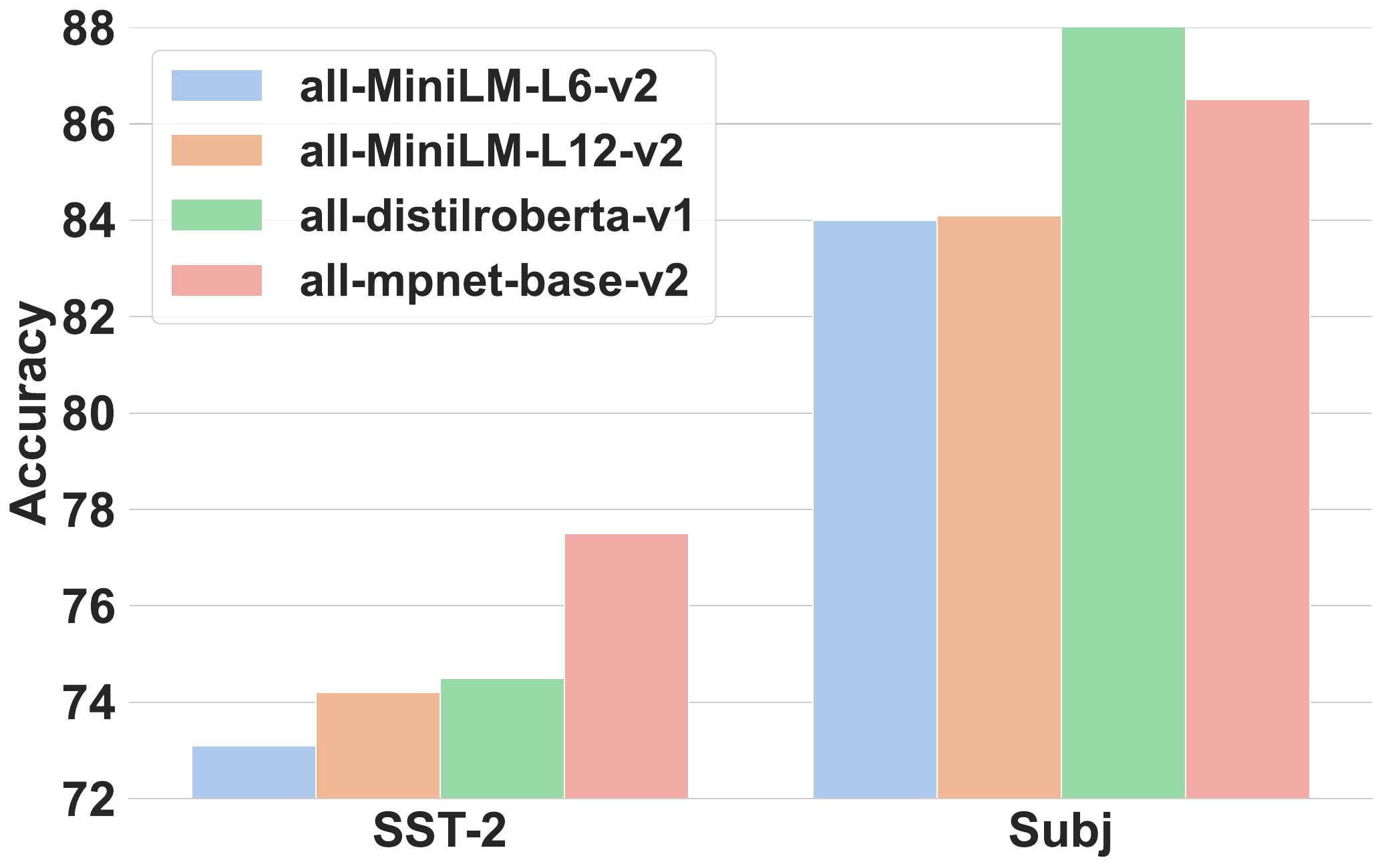}
    \caption{\textbf{The 1-shot performance with different retrieval models} on two classification datasets.}
    \label{fig:rtr}
\end{figure}
\subsection{Influencing Factors}
\label{subsec:influence_factors}
\paragraph{Preliminaries.} \textcolor{black}{The retrieval-based in-context learning paradigm primarily comprises four key components: demonstrations, test samples, the retrieval model and the inference model~\cite{xu2024context}. Previous extensive work~\cite{min-etal-2022-rethinking, liu2022makes, Selective_Annotation} has found that ICL performance is significantly influenced by the test data, and opting for test-similar demonstrations typically leads to yield superior performance. Although the effect of test data has been widely investigated, the model's impact has hardly been mentioned. To determine the influence of models, we proceed from both the retrieval model and inference model perspectives.}
\paragraph{Impact of Retrieval Models.} We first conduct experiments on classification tasks with different retrieval models. Specifically, we conduct experiments on two classification tasks, SST-2 and Subj~\cite{wang-etal-2018-glue}, with four sentence-transformer~\cite{reimers-gurevych-2019-sentence} models, including \textit{all-MiniLM-L6-v2}, \textit{all-MiniLM-L12-v2}, \textit{all-distilroberta-v1} and \textit{all-mpnet-base-v2}. As shown in Figure~\ref{fig:rtr}, the performance varies with different retrieval models and different datasets have different best retrievers. We speculate that the variance in model performance primarily arises from distinctions in similarity judgment between the retrieval model and the inference model. A smaller disparity in similarity judgment is expected to result in better in-domain demonstrations, which can improve the ICL performance~\cite{moore-lewis-2010-intelligent,sia-duh-2023-context}.

\begin{figure*}[t!]
    \centering
    \includegraphics[width=1.0\textwidth]{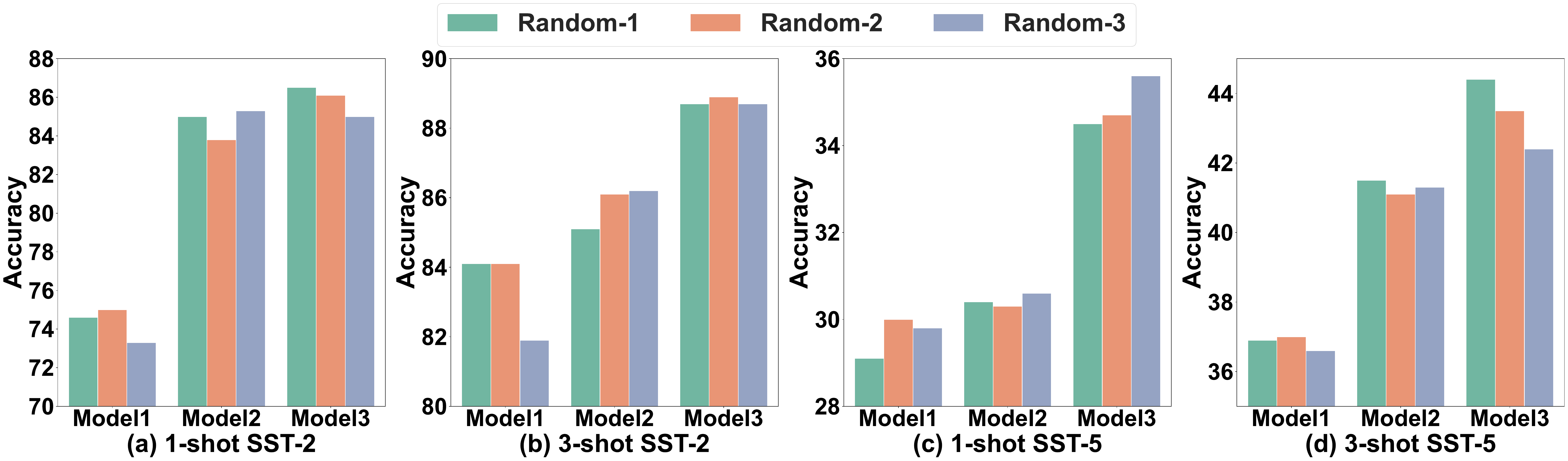}
    \caption{\textbf{The performance of different inference models with three randomly sampled demonstrations} for SST-2 and SST-5 datasets. Model1, Model2, Model3 represent GPT-J-6B, LLAMA2-7B, and LLAMA2-13B, respectively. The impact of various demonstrations varies depending on the specific inference models.}
    \label{fig:model_dep}
\end{figure*}

\paragraph{Impact of Inference Models.} 
The inference model is another factor that may influence the performance of in-context learning. 
To explore this, we conducted experiments on two classification tasks (e.g., SST-2 and SST-5) employing different inference models in both 1-shot and 3-shot settings. Specifically, we randomly sample different demonstrations 3 times for each test sample and assign them to Random-1, -2, and -3, respectively, and then we assess their performance across various inference models.
Results on Figure~\ref{fig:model_dep} show that the best demonstration varies across different inference models. For example, the performance of \textit{Random-2} is better than \textit{Random-3} in 1-shot SST-2 setting under \textit{llama2-7b} model, while the situation is totally reversed with \textit{llama2-13b}. We can also notice the same phenomenon under 3-shot settings, which implies increasing the in-context examples can not eliminate the influence of inference models. \textbf{Results above show that the choice of demonstrations is model-dependent.}

\subsection{Conjecture}
\label{subsec:assump}
Based on the above observations, we find demonstration choice is both data-dependent and model-dependent. Furthermore, \citet{gonen-etal-2023-demystifying} reveal that the more familiar the model is with prompts, the better the performance of prompts. Inspired by them, we propose a conjecture that \textit{effective demonstrations are those that can help the inference model better understand the test input.}

\textcolor{black}{To verify our assumption, we explore the relationship between the model's understanding of the test inputs and ICL performance. We simply employ the straightforward \textit{span shuffle} noise, which first selects sequences consisting of three consecutive tokens, and then randomly change their order,
following ~\citet{ding2022improving} to increase the difficulty of test input. Specifically, we first adopt TopK~\cite{liu2022makes} method to select the most test-similar demonstrations and compare the ICL performance of noised test samples with their original version.
Since the partial word shuffle will not influence people's reading~\cite{schad2012zoom,ward2013relationship}, our operation will not largely change the sentence's meaning. Table~\ref{tab:test_input} lists the results. We can notice that increasing the test samples' difficulty will lead to a large drop in ICL performance under both 1- and 3-shot settings, which reveals that
\textbf{ICL performance positively correlates with the model’s understanding of the test samples.}}

\begin{table}[ht]
\centering
\scalebox{0.78}{
\begin{tabular}{lccccccc}
\toprule
\multicolumn{1}{c}{\multirow{2}{*}{\bf Method}} &\multicolumn{3}{c}{\bf 1-shot} &\multicolumn{3}{c}{\bf 3-shot} \\ 
\cmidrule(lr){2-4} \cmidrule(lr){5-7} 
\multicolumn{1}{c}{} & \textit{SST-2} & \textit{SST-5} & \textit{Subj} & \textit{SST-2} & \textit{SST-5} & \textit{Subj}  \\ \midrule
Baseline & 81.9 & 38.1 & 89.8 & 79.2 & 39.0 &87.6 \\
\textit{shuffle} & 52.7 & 22.9 & 54.3 & 52.6 & 22.2 & 55.0  \\ \hdashline
\multicolumn{1}{c}{$\Delta$ ($\downarrow$)} & \textcolor{red}{\textbf{-29.2}} & \textcolor{red}{\textbf{-15.2}} & \textcolor{red}{\textbf{-35.5}} & \textcolor{red}{\textbf{-26.6}} & \textcolor{red}{\textbf{-16.8}} & \textcolor{red}{\textbf{-32.6}}  \\
\bottomrule
\end{tabular}
}
\caption{\textbf{Comparative results of GPT2-XL with origin test input and shuffled test input} on several tasks. We observe that the difficulty of test input will largely influence the ICL performance among all these tasks.}
\label{tab:test_input}
\end{table}

\section{Method}
Based on the above conclusions, we propose a simple and effective data- and model-dependent demonstration selection method, named \textbf{TopK + ConE}. Our method is based on the conjecture \textcolor{black}{in section \S\ref{subsec:assump},}
\textcolor{black}{which implies effective} demonstrations excel in reducing the conditional entropy of the test input under the inference model. 
It is noteworthy that we compute the conditional entropy of the test input rather than labels.
Mathematically, we find the best demonstrations $c^*$ by solving the following optimization problem: 
\begin{equation}
\label{eq:con_entropy}
    c^* = \argmin_{c \in \mathbf{C}} H_{\theta}(x|c),
\end{equation}
where each $c$ represents one possible demonstration group, and $H_{\theta}(x|c)$ signifies the inference model's uncertainty regarding the test input $x$ given the demonstrations $c$, which indicates the degree of the understanding of test input by the inference model. The lower the $H_{\theta}(x|c)$ is, the better the understanding is. The equation \eqref{eq:con_entropy} can be reformulated as
\begin{equation}
    c^* = \argmin_{c \in \mathbf{C}} (H_{\theta}(x, c) - H_{\theta}(c)),
\end{equation}
where $H_{\theta}(x, c)$ and $H_{\theta}(c)$ are the cross entropy of the whole prompt (including the demonstrations and test input) and the demonstrations estimated by the inference model, respectively. In other words, we are searching for demonstrations that minimize the difference of the cross-entropy between prompts and demonstrations. 

In the practical implementations, considering the huge search space generated by a large number of combinations, enumerating all combinations is infeasible. We adopt the selection-rerank framework proposed in \citet{wu-etal-2023-self}. Specifically, we first use the selection module to select the candidate demonstrations and then use our method to rank each candidate to get effective demonstrations.

\section{Experimental Setup}
\label{sec: exp}
\paragraph{Models.} We perform experiments across different sizes of models, including GPT2-XL (1.5B)~\cite{radford2019language}, GPT-j-6b (6B)~\cite{gpt-j}, Llama2-7b (7B) and Llama2-13b (13B)~\cite{touvron2023llama}, which are decoder-only dense LMs. We also conduct experiments on extensive alignment models, e.g., Llama2-7b-chat and Llama2-13b-chat~\cite{touvron2023llama}, Vicuna-7b, Vicuna-13b and Deepseek-7b-chat~\cite{deepseek-llm} to verify the generalizability of our approach.

\paragraph{Datasets.} We conduct a systematic study across 7 natural language understanding (NLU) tasks, including binary, multi-class classification tasks (SST-2, SST-5~\cite{socher-etal-2013-recursive}, CR, Subj~\cite{wang-etal-2018-glue}) and natural language inference tasks: MNLI~\cite{williams-etal-2018-broad} and QNLI~\cite{wang-etal-2018-glue}. We also evaluate our method in 4 machine translation tasks, extracted from Flores-200~\cite{goyal-etal-2022-flores} dataset, which contains 1012 examples for each language.

\paragraph{Baselines.} We mainly compare our method with five widely used methods that do not require additional training. 
\begin{itemize}
  \item \textbf{Prompting} is a special case of ICL without in-context examples.
  \item \textbf{Random} baseline randomly select in context examples for each testing sample.
  \item \textbf{BM25}~\cite{robertson2009probabilistic} baseline uses BM25 to calculate the word-overlap similarity between samples and test input, and select the high similarity samples as demonstrations.
  \item \textbf{TopK}~\cite{liu2022makes} baseline uses the nearest neighbors of a given test sample as the corresponding in-context examples.
  \item \textbf{TopK + MDL}~\cite{wu-etal-2023-self} adopt a select-then-rank framework, and rank the demonstrations selected by the TopK method based on the Minimum Description Length (MDL) principle.
\end{itemize}

\paragraph{Evaluation Metrics.} We adopt different evaluation methods for different tasks. For classification, we report the performance with the Accuracy. For the translation tasks, we adopt the mostly used language model-based metrics \textbf{COMET}~\cite{rei-etal-2020-comet} since they have demonstrated a high correlation with human evaluation and are resilient to domain shift. Specifically, we use the reference-based metric COMET-20 (\textit{wmt20-COMET-da}) and COMET-22 (\textit{wmt22-COMET-da}) for evaluation, and use the default parameters of "comet-compare" for the significance test\footnote{\url{https://github.com/Unbabel/COMET}}.

\paragraph{Experimental Details.} We use the TopK method to retrieve 30 candidates for each sample, and then rank each candidate using our ConE method. Templates are adopted from~\citet{lu-etal-2022-fantastically, wu-etal-2023-self} and detailed in Table~\ref{tab:templates}. We ran all experiments 3 times with different random seeds and reported the average accuracies. We use 4-shot ICL for GPT2-XL and 8-shot for others, the ablations are in \S\ref{sec:ab}. Our codebase is built based on OpenICL~\cite{wu-etal-2023-openicl}. 

\begin{table*}
\centering
\resizebox{0.9\linewidth}{!}{
\begin{tabular}{lcccccccl} 
\toprule
                            \textbf{Method}     & \textbf{SST-2} & \textbf{CR}   & \textbf{Subj} & \textbf{SST-5} & \textbf{AGNews}      & \textbf{MNLI} & \textbf{QNLI} &  \textit{\textbf{Average}}                                         \\ 
\midrule
\textbf{Prompting}                        & 68.7           & 81.1          & 49.4          & 25.3           & 67.0                             & 47.5          & 53.3          & 56.0 (+21.9)                                        \\
\hdashline
\textbf{Random}                           & 94.4           & 92.3          & 70.9          & 50.4           & 83.5                               & 51.0          & 56.2         & 71.2 (+6.8)                                        \\
\textbf{BM25}                             & 94.5           & 92.8          & 76.8          & 52.6           & 92.5                               & 57.0          & 59.0         & 75.0 (+2.9)                                        \\
\textbf{TopK}                             & 95.2           & 92.8          & 80.4          & 52.6           & 92.4                              & 57.8          & 61.3          & 76.1 (+1.8)                                        \\
\textbf{TopK + MDL}                       & 95.1           & \textbf{93.4} & 81.2          & \textbf{52.7}  & 92.3                              & 57.9          & 64.5          & 76.7 (+1.2)                                        \\ 
\midrule
\textbf{Ours}                             & \textbf{95.4}  & 93.1          & \textbf{85.8} & 52.5           & \textbf{92.8}             & \textbf{59.5} & \textbf{66.4} & \textbf{77.9}                           \\
\bottomrule
\end{tabular}}
\caption{\textbf{Performance of different methods across 7 Natural Language Understanding (NLU) tasks} on Llama2-7B model. The best results are in \textbf{bold}. We can see that our method improves the performance of almost all task types. Numbers in the parenthesis represent the relative improvements our method achieved over baselines.}
\label{tab:nlu_tasks}
\end{table*}

\section{Main Results}
\subsection{Natural Language Understanding Tasks}
\begin{figure}[t!]
    \centering
    \includegraphics[width=0.9\columnwidth]{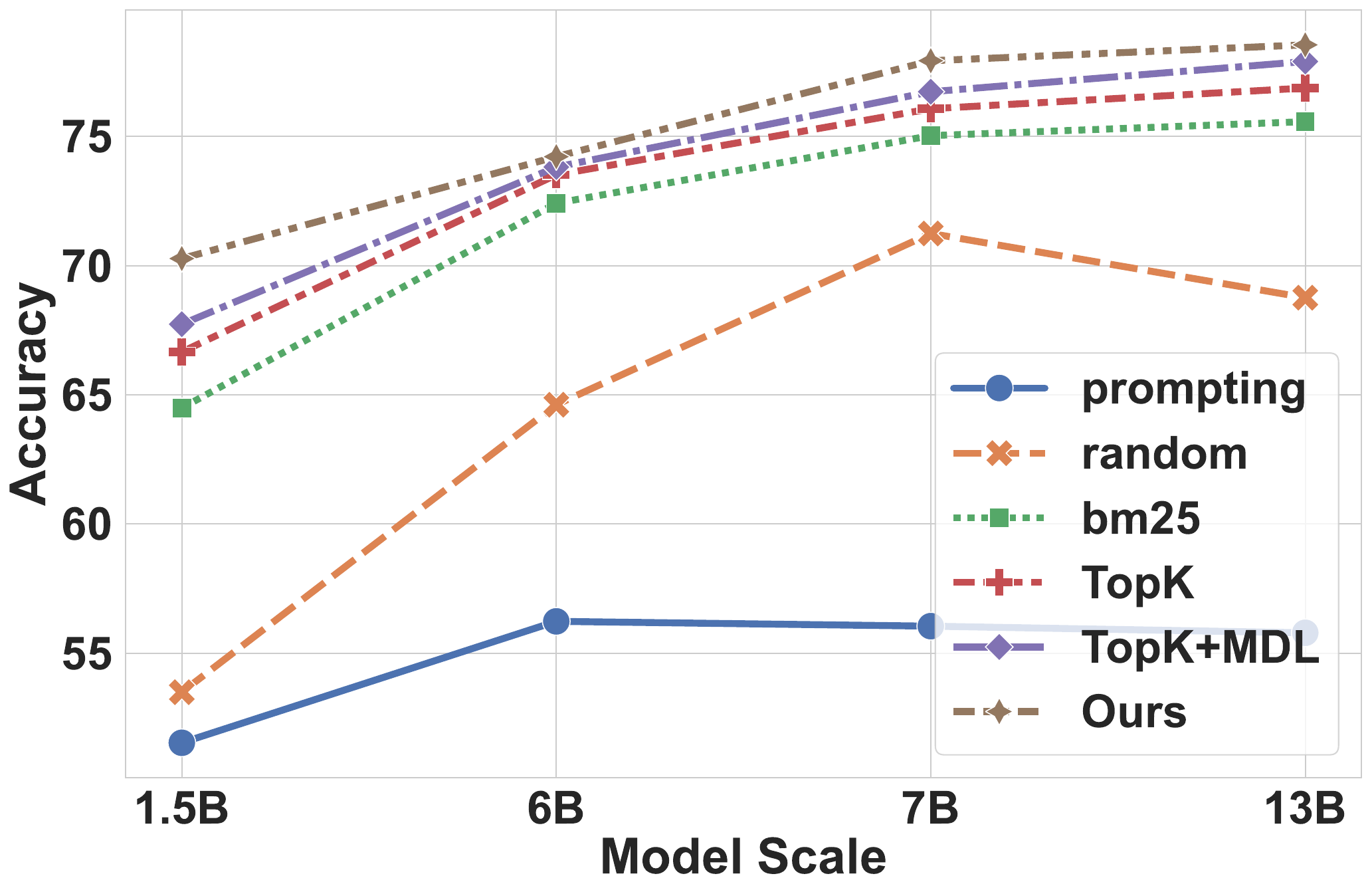}
    \caption{\textbf{The average performance of 7 NLU tasks across different model scales.} Our method consistently outperforms previous methods across model scales.}
    \label{fig:model_nlu}
\end{figure}

We first verify the effectiveness of our method in NLU Tasks. Specifically, we conduct experiments on 7 classification tasks, including binary classification tasks, multi-class classification tasks, and natural language inference tasks. Based on the results on Table~\ref{tab:nlu_tasks} and Figure~\ref{fig:model_nlu}, we can find that:

\paragraph{Our method brings consistent performance improvements on almost all types of tasks.} Results in Table~\ref{tab:nlu_tasks} show the superior performance of our approach compared to the existing state-of-the-art method, \textit{TopK+MDL}, across the majority of tasks, resulting in an average accuracy improvement of 1.2\%. Compared with our selection method \textit{TopK}, our method considerably improves the performance on 6 tasks out of the total 7 tasks, yielding an average gain of 1.8\%, proving the effectiveness of improving the model's understanding to test input. Furthermore, it is noteworthy that our approach can achieve significant improvements in challenging tasks, such as the Subj and QNLI tasks, respectively bringing 4.6\% and 1.9\% gains compared to the previously optimal methods, demonstrating the superior performance of our method for hard-to-understanding tasks.

\paragraph{Our method brings gains across different model sizes.} Figure~\ref{fig:model_nlu} presents the average performance across 7 Natural Language Understanding (NLU) tasks using various inference models, ranging in size from 1.5B (\textit{GPT2-XL}) to 13B (\textit{Llama2-13B}). Results reveal that advanced ICL methods usually can achieve better performance when we scale up the model size, while prompting and random ICL methods will produce unstable results. Notably, our approach consistently outperforms previous methods across different model scales, particularly in the case of GPT2-XL, which yields an average gain of 2.6\% and 3.6\% compared to \textit{TopK+MDL} and \textit{TopK} methods. 

\begin{table*}
\centering
\resizebox{0.9\linewidth}{!}{
\begin{tabular}{ccccccccc} 
\toprule
\multirow{2}{*}{\textbf{Method}} & \multicolumn{2}{c}{\textbf{En$\Rightarrow$Zh}}  & \multicolumn{2}{c}{\textbf{Zh$\Rightarrow$En}}  & \multicolumn{2}{c}{\textbf{Ru$\Rightarrow$De}}  & \multicolumn{2}{c}{\textbf{De$\Rightarrow$Ru}}   \\ 
\cmidrule(lr){2-3}\cmidrule(r){4-5}\cmidrule(lr){6-7}\cmidrule(r){8-9}
                                 & \textbf{COMET20} & \textbf{COMET22} & \textbf{COMET20} & \textbf{COMET22} & \textbf{COMET20} & \textbf{COMET22} & \textbf{COMET20} & \textbf{COMET22}  \\ 
\midrule
\multicolumn{9}{c}{\textit{-w/ 1-shot}}                                                                                                                                                   \\ 
\hline\hline
Random                           & 35.7             & 81.5             & 60.9             & 85.1             & 44.0             & 79.8             & \textbf{52.4}    & 83.6              \\
BM25                             & 35.1             & 81.3             & 60.9             & 85.1             & 42.2             & 79.5             & 50.2             & 83.4              \\
TopK                             & 35.9             & 81.5             & 61.0             & 85.1             & \textbf{43.9}             & 79.7             & 49.7             & 83.3              \\ 
\hdashline
Ours                             & \textbf{37.1$^\dagger$}    & \textbf{81.7$^\dagger$}    & \textbf{61.7$^\dagger$}    & \textbf{85.4$^\dagger$}    & \textbf{43.9}    & \textbf{79.9}    & 51.8$^\dagger$             & \textbf{83.8$^\dagger$}     \\ 
\midrule
\multicolumn{9}{c}{\textit{-w/ 3-shot}}                                                                                                                                                   \\ 
\hline\hline
Random                           & 40.1             & 82.4             & 62.7             & 85.5             & 47.8             & 80.6             & 54.6             & 84.0              \\
BM25                             & 39.6             & 82.3             & 62.3             & 85.4             & 47.0             & 80.5             & 53.2             & 83.9              \\
TopK                             & 39.9             & 82.4             & \textbf{63.3}             & 85.6             & 46.8             & 80.4             & 53.1             & 83.9              \\ 
\hdashline
Ours                             & \textbf{40.7$^\dagger$}    & \textbf{82.6$^\dagger$}    & \textbf{63.3}    & \textbf{85.7}    & \textbf{47.9$^\dagger$}    & \textbf{80.8$^\dagger$}    & \textbf{55.3$^\dagger$}    & \textbf{84.5$^\dagger$}     \\
\bottomrule
\end{tabular}}
\caption{\textbf{Performance on different methods across 4 language pairs on Llama2-7b}. The best results are in \textbf{bold}. ``$^\dagger$'' indicates a statistically significant difference from the TopK baseline ($p<0.05$).}
\label{tab:nlg}
\end{table*}

\subsection{Natural Language Generation Tasks}
We further evaluate our method on NLG tasks, i.e. machine translation. Recent study~\cite{gpt2023} reveals that LLMs have achieved comparable or better performance on par with their best-supervised counterpart systems~\cite{zan2022vega} in competing WMT Chinese-English tasks. We conduct experiments in 4 language pairs, including English-centric language pairs and non-English-centric language pairs. 

\paragraph{Results.} The results across different language pairs under different settings are presented in Table~\ref{tab:nlg}.
Obviously, our method can consistently improve the performance of ICL in terms of COMET score compared with TopK in both English-centric and non-English-centric language pairs. Especially in non-English-centric language pairs, our method brings +1.1 and +2,2 COMET20 score improvement in Ru$\Rightarrow$De and De$\Rightarrow$Ru under the 3-shot setting, respectively. We attribute this to the improvement of the model's understanding of the test sample, and the more difficult the sample, the greater the benefit from our method. 
Furthermore, we can notice that previous advanced ICL methods do not always work, especially for non-English centric language pairs, while our method can consistently achieve the best performance under the 3-shot settings, demonstrating the effectiveness of our method on generation tasks.

\section{Analysis}
To further demonstrate the effectiveness and generality of our method, we conduct further analyses on NLU tasks (with the GPT2-XL model) and NLG tasks (with Llama2-7b).

\begin{figure*}[t!]
    \centering
    \includegraphics[width=0.9\linewidth]{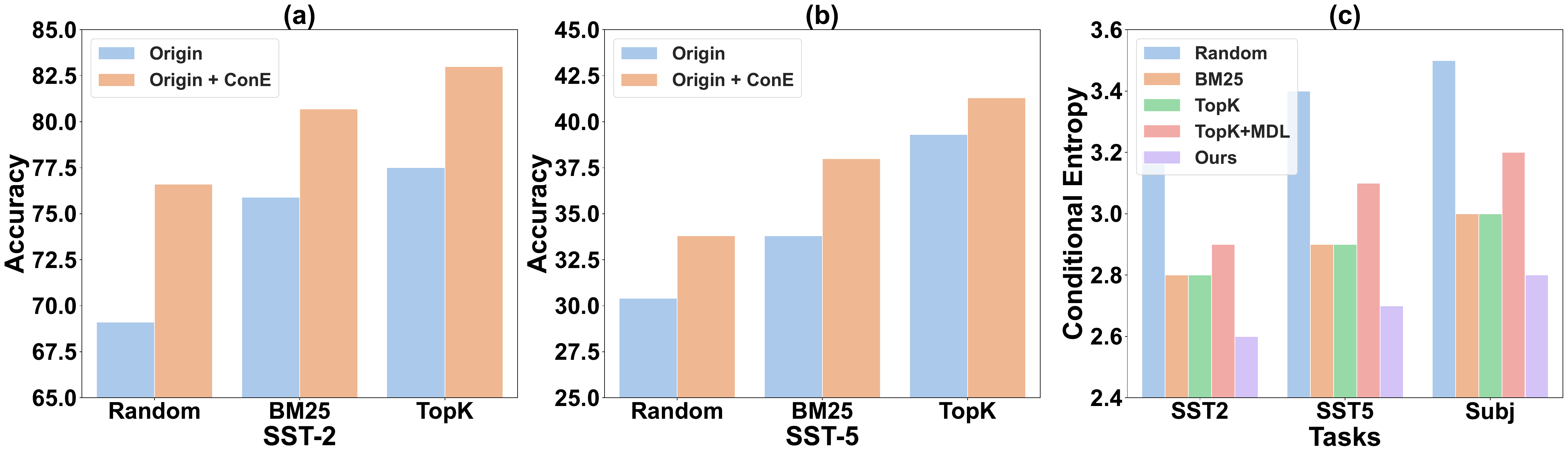}
    \caption{(a, b) \textbf{The effect of our method with different selection methods} in SST-2 and SST-5, \textbf{origin} represents the baseline method without our ConE method, while \textbf{Origin + ConE} signifies with our ConE method. (c) \textbf{The conditional entropy of the test input} with different ICL methods. }
    \label{fig: analysis}
\end{figure*}

\paragraph{Our ConE method is complementary to previous approaches.} To further explore the generality of our method, we combine ConE with different selection methods, e.g. random and BM25, in binary and multi-choice classification tasks. The results in Figure~\ref{fig: analysis} (a, b) show that ConE can further significantly improve the baseline performance in different types of tasks. Especially in SST-2 tasks with the Random method, ConE brings +7.5 score improvement, which indicates that ConE is complementary to previous approaches and can further improve their performance. We can also notice that TopK + ConE achieves better performance compared with other methods, hence we choose TopK as our selection method because of its simplicity and effectiveness.

\paragraph{Our method works for mix-domain demonstration pools.} Previous results have shown the superior performance of our method in single-domain demonstration pools. Now, we evaluate the effectiveness of our method in mixed demonstration pools, which have demonstrations from different domains. Specifically, we evaluate the performance of our method in three domains, e.g., e-commerce, news and social, with a mix-domain demonstrations pool in WMT22 translation task\footnote{\url{https://www.statmt.org/wmt22/translation-task.html}}. Experimental results in Table~\ref{tab:mix_domain} show that our method can achieve consistent improvements in three domains with 3-shot ICL, especially in Zh$\Rightarrow$En, which achieve over 1.0 COMET improvement across three domains, showing that \textit{our method also works for mix-domain demonstration pools.}

\begin{table}
\centering
\resizebox{1\linewidth}{!}{
\begin{tabular}{ccccccc} 
\toprule
\multirow{2}{*}{\textbf{Method}} & \multicolumn{3}{c}{\textbf{Zh$\Rightarrow$En}}                    & \multicolumn{3}{c}{\textbf{En$\Rightarrow$Zh}}                     \\ 
\cmidrule(lr){2-4}\cmidrule(lr){5-7}
                                 & \textbf{ecommerce} & \textbf{news} & \textbf{social} & \textbf{ecommerce} & \textbf{news} & \textbf{social}  \\ 
\midrule
\multicolumn{7}{c}{\textit{-w/ 1-shot}}                                                                                                         \\ 
\hline\hline
random                           & 3.7                & 29.8          & 31.2            & 33.0               & 17.8          & \textbf{6.8}     \\
TopK                             & \textbf{6.4}       & 30.6          & \textbf{32.4}   & 32.6               & 18.1          & 6.1              \\ 
\hdashline
Ours                             & 6.0                & \textbf{33.2$^\dagger$} & \textbf{32.4}   & \textbf{36.1$^\dagger$}      & \textbf{21.0$^\dagger$} & 4.8              \\ 
\midrule
\multicolumn{7}{c}{\textit{-w/ 3-shot}}                                                                                                         \\ 
\hline\hline
random                           & 8.1                & 33.3          & 33.4            & 34.3               & 22.4          & 11.3             \\
TopK                             & 7.5                & 35.3          & 33.3            & 36.7               & 24.1          & 11.9             \\ 
\hdashline
Ours                             & \textbf{9.5$^\dagger$}       & \textbf{36.3$^\dagger$} & \textbf{34.4$^\dagger$}   & \textbf{37.0}      & \textbf{25.2$^\dagger$} & \textbf{12.5$^\dagger$}    \\
\bottomrule
\end{tabular}}
\caption{\textbf{Performance of our method for domain dataset with a mixed-domain demonstration pool} with inference model Llama2-7b. ``$^\dagger$'' indicates a statistically significant difference from the TopK baseline ($p<0.05$).}
\label{tab:mix_domain}
\end{table}

\paragraph{Our method works for aligned chat models.}
\begin{figure}[t!]
    \centering
    \includegraphics[width=1.0\linewidth]{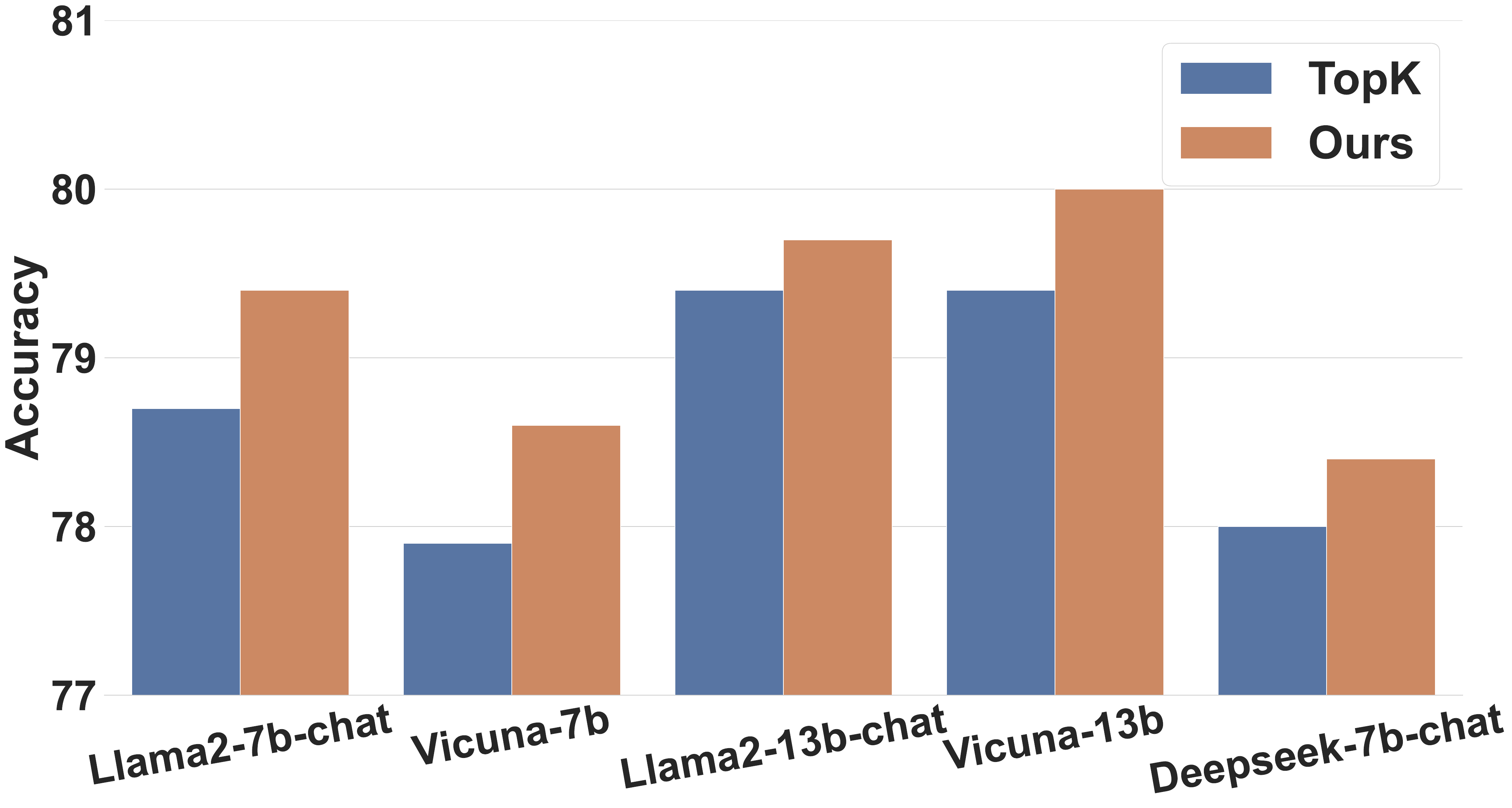}
    \caption{\textbf{The average performance of different chat models} in 7 NLU tasks.}
    \label{fig:sft_model}
\end{figure}
To verify the effectiveness of our method for the chat LLMs, we conducted extensive experiments on different instruction-tuned and RLHF-tuned LLMs, including Vicuna, LLaMA-chat, and DeepSeek-chat. The results in Figure~\ref{fig:sft_model} show that our method can achieve consistent improvement in different models, demonstrating that \textit{our method also works for instruction-tuned and safety-enhanced models}.

\section{Impact of hyperparameter}
\label{sec:ab}
\begin{figure*}[t!]
    \centering
    \includegraphics[width=0.9\linewidth]{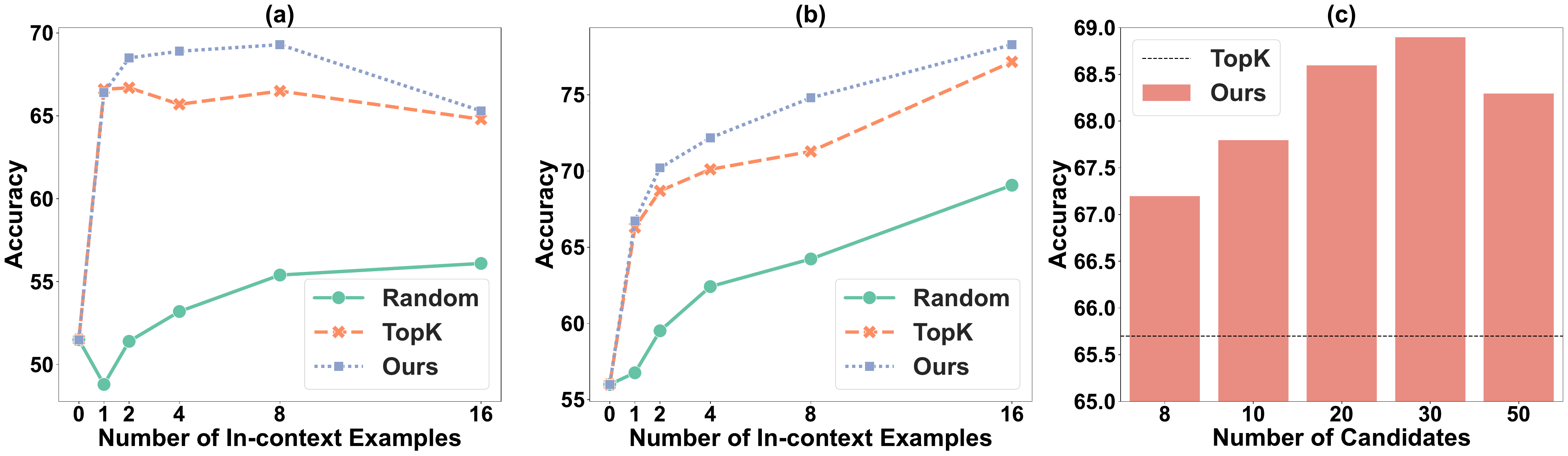}
    \caption{\textbf{The average performance of ablation experiments}. (a, b) Impact of the number of in-context examples for GPT2-XL and Llama2-7b; (c) Impact of the number of candidates selected by the TopK method.}
    \label{fig:ablation}
\end{figure*}
In this section, we conduct ablation studies on the hyperparameters in our method. 
\paragraph{Impact of In-context Examples.} We gradually increase the number of in-context examples (denoted as N) from 0 (prompting) to 16. The results are listed in Figure~\ref{fig:ablation}(a, b), we see that increasing N
usually can consistently improve the performance on average, but when N=16 the ICL performance in GPT2-XL degrades.
Through further analysis, we found that the decrease comes from the constraint of the maximum sentence length of the model (GPT2-XL), and the phenomenon even occurs when we set N as 8 for GPT2-XL. Hence, we choose N=4 for GPT2-XL, and N=8 for other models. Note that our method can consistently outperform the TopK method, and increasing the in-context examples can further improve our method.

\paragraph{Impact of Candidate Numbers.} As mentioned above, our method comprises two modules: the TopK selection and the ConE reranking. The selection module will reduce the space of in-context examples to speed up the whole process. Hence we explore the impact of the candidate numbers selected by TopK. The results in Figure~\ref{fig:ablation}(c) list the performance of 4 in-context examples with the GPT2-XL model. We can notice that our method is always better than the baseline TopK, and increasing the number of candidates can further improve the performance. Based on the results, we set the default candidate number as 30. 

\section{Discussion}
\paragraph{Whether our method can partially explain why previous ICL methods work?}
Intuitively, enhancing the model's understanding to test input is one of the reasons why previous methods work. To prove this, 
we calculate the conditional entropy of the test input with respect to previous baselines across three classification tasks. The results presented in Figure~\ref{fig: analysis}(c) show that the previous methods will also reduce the conditional entropy of test samples in all three tasks, which demonstrate that previous ICL methods can also be explained by our conjecture. These results show the universality of our conjecture.

\paragraph{Whether our method is sensitive to the demonstration order?}
Previous studies have proven that ICL is very sensitive to the order of in-context examples~\cite{lu-etal-2022-fantastically}. To explore the sensitivity of our methods for the order of in-context examples, we randomize the order of our chosen demonstrations on three classification tasks and compare the stability with Random and TopK methods. Results on Table~\ref{tab:order} show that our method can achieve better average performance with smaller variance among all tasks, demonstrating that \textit{our method could alleviate the order sensitivity issue in the ICL framework.}

\begin{table}
\centering
\resizebox{1\linewidth}{!}{
\begin{tabular}{ccccccc} 
\toprule
\multirow{2}{*}{\textbf{Method}} & \multicolumn{2}{c}{\textbf{SST-2}}      & \multicolumn{2}{c}{\textbf{Subj}}       & \multicolumn{2}{c}{\textbf{CR}}          \\ 
\cmidrule(lr){2-3}\cmidrule(lr){4-5}\cmidrule(lr){6-7}
                                 & \textbf{\textit{Avg.}} & \textbf{\textit{Var.}}           & \textbf{\textit{Avg.}} & \textbf{\textit{Var.}}            & \textbf{\textit{Avg.}} & \textbf{\textit{Var.}}             \\ 
\midrule
\textbf{Random}                  & 68.8          & 0.90                   & 56.7          & 0.49                    &  67.8         & 4.00                     \\
\textbf{TopK}                    & 78.6          & 0.56                    & 86.2          & 0.20                    & 73.9          & 0.61                     \\ 
\hdashline
\textbf{Ours}                    & \textbf{82.0} & \textbf{0.26} & \textbf{91.0} & \textbf{0.05} & \textbf{81.0} & \textbf{0.26}  \\
\bottomrule
\end{tabular}}
\caption{\textbf{The average performance and variance of 10 random permutations of four in-context examples} for GPT2-XL. The best results are in \textbf{bold}. \textit{Our method achieves consistently better average performance with lower variance.}}
\label{tab:order}
\end{table}

\section{Related Work}
Despite that large language models have shown their surprising zero-shot performance in various tasks, even including complex reason/ agent tasks~\cite{wang2024oop,ren2024healthcare,zhang2024intention,zhong2024achieving}. Recent works show that ICL can effectively elicit their capability and further improve LLMs' performance~\cite{dong2022survey}. Besides effectiveness, ICL may provide format guidance to alleviate the prompt bias during language model inference~\cite{xu-etal-2024-take-care}.
However, the performance of ICL is unstable~\cite{lu-etal-2022-fantastically}, and the small change of in-context examples and their order can result in a large variance in performance. 

Motivated by the instability of the ICL performance, in-context example selection methods have been widely investigated. ~\citet{lu-etal-2022-fantastically} first propose a validation-free corpus-level method for determining the optimal order of in-context examples. However, they only investigate the influence of order without proposing how to better select in-context examples. Inspired by the success of retrieval modules in augmenting neural networks, ~\citet{liu2022makes} find examples that are close to each test sample in embedding space that can serve as a good choice for ICL. Following the finding of ~\citet{liu2022makes}, ~\citet{Selective_Annotation} subsequently extended their method by incorporating increased diversity in the selection of in-context examples. However, why these methods work is still unclear and the methods only consider the influence from the data aspect. 

Unlike the data-dependent demonstration selection methods, model-dependent methods are rarely explored. ~\citet{wu-etal-2023-self} proposed a demonstration rank method grounded in the minimum description length principle, which utilizes the inference model to select the optimal in-context example organization. However, their ranked in-context organizations are randomly sampled, which may limit their performance. ~\citet{wang-etal-2024-learning} proposed a novel framework to iteratively train dense retrievers to identify high-quality in-context examples for LLMs. However, they need additional training, which is costly for practitioners. Furthermore, both methods neglected to investigate whether and how the inference model affects ICL performance.

On the other hand, although some previous methods~\cite{wu-etal-2023-self, iter-etal-2023-context, wang-etal-2023-label} have emphasized the significance of understanding the test samples, their primary emphasis lies in the confidence of test labels, neglecting that of test input. For instance, ~\citet{wu-etal-2023-self} searches the demonstrations capable of losslessly compressing testing labels, and ~\citet{iter-etal-2023-context} identify the in-domain demonstrations through the cross-entropy difference of test labels computed by the small model fine-tuned in demonstrations. While ~\citet{wang-etal-2023-label} propose to reweight label anchors to improve ICL performance. ~\citet{gonen-etal-2023-demystifying} found that using perplexity could be a good heuristic for prompt selection, while the effect for ICL has not been investigated.

\section{Conclusion}
In this paper, we take the first step to investigate the factors that influence the choice of demonstrations in ICL from the model perspective, and find that the demonstration selection is both data- and model-dependent. 
Based on the findings, we conjecture that effective demonstrations can improve the inference model’s understanding to test input, and correspondingly propose a data- and model-dependent selection method. 
Empirical results suggest that our method can significantly outperform the previous ICL method. Further analysis confirms the generalization of our method and our approach can provide a unified explanation for previous studies.

\section*{Limitations}
Our work has several potential limitations.
First, given the limited computational budget, we only validate our \textbf{TopK + ConE} on the 1.5B-13B LLMs. It will make our work more convincing if scaling the experiments up to the larger model size, \textit{e.g.}, 70B. On the other hand, our method introduces some computational budgets during the inference for select demonstrations, which may be unacceptable for extremely large LLMs. It is meaningful to explore a more efficient method to measure the model's understanding to test input to accelerate the process of demonstration selection, which is in our future work.

\section*{Ethic Statements}
We take ethical considerations very seriously and strictly adhere to the ACL Ethics Policy. This paper focuses on the in-context learning behaviour of LLMs and proposes a data- and model-dependent demonstration selection method to improve ICL performance. To explore the influencing factors of ICL, we revisit the demonstration selection Strategies from model aspect, and propose a conjecture to find effective demonstrations.
However, it should be noted that all pretrained models and evaluation datasets used in this study are publicly available and have been widely adopted by researchers. We do not proactively introduce additional data or models that may cause ethical issues, and we believe that our proposed method will help alleviate ethical issues.

\section*{Acknowledgments}
This work is supported by the National Natural Science Foundation of China (No. 62377002). Xuebo Liu was sponsored by CCF-Tencent Rhino-Bird Open Research Fund. We are grateful to the anonymous reviewers and the area chair for their insightful comments and suggestions. 

\bibliography{custom}

\begin{thebibliography}{45}
\expandafter\ifx\csname natexlab\endcsname\relax\def\natexlab#1{#1}\fi

\bibitem[{Brown et~al.(2020{\natexlab{a}})Brown, Mann, Ryder, Subbiah, Kaplan, Dhariwal, Neelakantan, Shyam, Sastry, Askell et~al.}]{brown2020language}
Tom Brown, Benjamin Mann, Nick Ryder, Melanie Subbiah, Jared~D Kaplan, Prafulla Dhariwal, Arvind Neelakantan, Pranav Shyam, Girish Sastry, Amanda Askell, et~al. 2020{\natexlab{a}}.
\newblock \href {https://proceedings.neurips.cc/paper_files/paper/2020/file/1457c0d6bfcb4967418bfb8ac142f64a-Paper.pdf} {Language models are few-shot learners}.
\newblock \emph{NeurIPS}.

\bibitem[{Brown et~al.(2020{\natexlab{b}})Brown, Mann, Ryder, Subbiah, Kaplan, Dhariwal, Neelakantan, Shyam, Sastry, Askell, Agarwal, Herbert{-}Voss, Krueger, Henighan, Child, Ramesh, Ziegler, Wu, Winter, Hesse, Chen, Sigler, Litwin, Gray, Chess, Clark, Berner, McCandlish, Radford, Sutskever, and Amodei}]{DBLP:conf/nips/BrownMRSKDNSSAA20}
Tom~B. Brown, Benjamin Mann, Nick Ryder, Melanie Subbiah, Jared Kaplan, Prafulla Dhariwal, Arvind Neelakantan, Pranav Shyam, Girish Sastry, Amanda Askell, Sandhini Agarwal, Ariel Herbert{-}Voss, Gretchen Krueger, Tom Henighan, Rewon Child, Aditya Ramesh, Daniel~M. Ziegler, Jeffrey Wu, Clemens Winter, Christopher Hesse, Mark Chen, Eric Sigler, Mateusz Litwin, Scott Gray, Benjamin Chess, Jack Clark, Christopher Berner, Sam McCandlish, Alec Radford, Ilya Sutskever, and Dario Amodei. 2020{\natexlab{b}}.
\newblock \href {https://proceedings.neurips.cc/paper/2020/hash/1457c0d6bfcb4967418bfb8ac142f64a-Abstract.html} {Language models are few-shot learners}.
\newblock In \emph{NeurIPS}.

\bibitem[{DeepSeek-AI(2024)}]{deepseek-llm}
DeepSeek-AI. 2024.
\newblock \href {https://github.com/deepseek-ai/DeepSeek-LLM} {Deepseek llm: Scaling open-source language models with longtermism}.
\newblock \emph{arXiv preprint arXiv:2401.02954}.

\bibitem[{Ding et~al.(2022)Ding, Peng, and Tao}]{ding2022improving}
Liang Ding, Keqin Peng, and Dacheng Tao. 2022.
\newblock \href {https://arxiv.org/abs/2201.07365} {Improving neural machine translation by denoising training}.
\newblock \emph{arXiv preprint}.

\bibitem[{Dong et~al.(2022)Dong, Li, Dai, Zheng, Wu, Chang, Sun, Xu, and Sui}]{dong2022survey}
Qingxiu Dong, Lei Li, Damai Dai, Ce~Zheng, Zhiyong Wu, Baobao Chang, Xu~Sun, Jingjing Xu, and Zhifang Sui. 2022.
\newblock \href {https://arxiv.org/abs/2301.00234} {A survey for in-context learning}.
\newblock \emph{arXiv preprint}.

\bibitem[{Gonen et~al.(2023)Gonen, Iyer, Blevins, Smith, and Zettlemoyer}]{gonen-etal-2023-demystifying}
Hila Gonen, Srini Iyer, Terra Blevins, Noah Smith, and Luke Zettlemoyer. 2023.
\newblock \href {https://aclanthology.org/2023.findings-emnlp.679} {Demystifying prompts in language models via perplexity estimation}.
\newblock In \emph{Findings of EMNLP}.

\bibitem[{Goyal et~al.(2022)Goyal, Gao, Chaudhary, Chen, Wenzek, Ju, Krishnan, Ranzato, Guzm{\'a}n, and Fan}]{goyal-etal-2022-flores}
Naman Goyal, Cynthia Gao, Vishrav Chaudhary, Peng-Jen Chen, Guillaume Wenzek, Da~Ju, Sanjana Krishnan, Marc{'}Aurelio Ranzato, Francisco Guzm{\'a}n, and Angela Fan. 2022.
\newblock \href {https://aclanthology.org/2022.tacl-1.30} {The {F}lores-101 evaluation benchmark for low-resource and multilingual machine translation}.
\newblock \emph{TACL}.

\bibitem[{Hashimoto et~al.(2018)Hashimoto, Guu, Oren, and Liang}]{hashimoto2018retrieve}
Tatsunori~B Hashimoto, Kelvin Guu, Yonatan Oren, and Percy~S Liang. 2018.
\newblock \href {https://proceedings.neurips.cc/paper_files/paper/2018/file/cd17d3ce3b64f227987cd92cd701cc58-Paper.pdf} {A retrieve-and-edit framework for predicting structured outputs}.
\newblock \emph{NeurIPS}.

\bibitem[{Hendy et~al.(2023)Hendy, Abdelrehim, Sharaf, Raunak, Gabr, Matsushita, Kim, Afify, and Awadalla}]{gpt2023}
Amr Hendy, Mohamed Abdelrehim, Amr Sharaf, Vikas Raunak, Mohamed Gabr, Hitokazu Matsushita, Young~Jin Kim, Mohamed Afify, and Hany~Hassan Awadalla. 2023.
\newblock \href {https://arxiv.org/abs/2302.09210} {How good are gpt models at machine translation? a comprehensive evaluation}.
\newblock \emph{arXiv preprint}.

\bibitem[{Iter et~al.(2023)Iter, Pryzant, Xu, Wang, Liu, Xu, and Zhu}]{iter-etal-2023-context}
Dan Iter, Reid Pryzant, Ruochen Xu, Shuohang Wang, Yang Liu, Yichong Xu, and Chenguang Zhu. 2023.
\newblock \href {https://aclanthology.org/2023.findings-emnlp.81} {In-context demonstration selection with cross entropy difference}.
\newblock In \emph{EMNLP}.

\bibitem[{Liu et~al.(2022)Liu, Shen, Zhang, Dolan, Carin, and Chen}]{liu2022makes}
Jiachang Liu, Dinghan Shen, Yizhe Zhang, William~B Dolan, Lawrence Carin, and Weizhu Chen. 2022.
\newblock \href {https://aclanthology.org/2022.deelio-1.10/} {What makes good in-context examples for gpt-3?}
\newblock In \emph{DeeLIO}.

\bibitem[{Lu et~al.(2023)Lu, Qiu, Ding, Zhang, Kocmi, and Tao}]{Lu2023EAPrompt}
Qingyu Lu, Baopu Qiu, Liang Ding, Kanjian Zhang, Tom Kocmi, and Dacheng Tao. 2023.
\newblock \href {https://arxiv.org/abs/2303.13809} {Error analysis prompting enables human-like translation evaluation in large language models: A case study on chatgpt}.
\newblock \emph{arXiv preprint}.

\bibitem[{Lu et~al.(2022)Lu, Bartolo, Moore, Riedel, and Stenetorp}]{lu-etal-2022-fantastically}
Yao Lu, Max Bartolo, Alastair Moore, Sebastian Riedel, and Pontus Stenetorp. 2022.
\newblock \href {https://aclanthology.org/2022.acl-long.556} {Fantastically ordered prompts and where to find them: Overcoming few-shot prompt order sensitivity}.
\newblock In \emph{ACL}.

\bibitem[{Min et~al.(2022)Min, Lyu, Holtzman, Artetxe, Lewis, Hajishirzi, and Zettlemoyer}]{min-etal-2022-rethinking}
Sewon Min, Xinxi Lyu, Ari Holtzman, Mikel Artetxe, Mike Lewis, Hannaneh Hajishirzi, and Luke Zettlemoyer. 2022.
\newblock \href {https://aclanthology.org/2022.emnlp-main.759} {Rethinking the role of demonstrations: What makes in-context learning work?}
\newblock In \emph{EMNLP}.

\bibitem[{Moore and Lewis(2010)}]{moore-lewis-2010-intelligent}
Robert~C. Moore and William Lewis. 2010.
\newblock \href {https://aclanthology.org/P10-2041} {Intelligent selection of language model training data}.
\newblock In \emph{ACL}.

\bibitem[{Ouyang et~al.(2022)Ouyang, Wu, Jiang, Almeida, Wainwright, Mishkin, Zhang, Agarwal, Slama, Ray et~al.}]{ouyang2022training}
Long Ouyang, Jeffrey Wu, Xu~Jiang, Diogo Almeida, Carroll Wainwright, Pamela Mishkin, Chong Zhang, Sandhini Agarwal, Katarina Slama, Alex Ray, et~al. 2022.
\newblock \href {https://proceedings.neurips.cc/paper_files/paper/2022/file/b1efde53be364a73914f58805a001731-Paper-Conference.pdf} {Training language models to follow instructions with human feedback}.
\newblock \emph{NeurIPS}.

\bibitem[{Peng et~al.(2023)Peng, Ding, Zhong, Shen, Liu, Zhang, Ouyang, and Tao}]{Peng2023ChatGPT4MT}
Keqin Peng, Liang Ding, Qihuang Zhong, Li~Shen, Xuebo Liu, Min Zhang, Yuanxin Ouyang, and Dacheng Tao. 2023.
\newblock \href {https://aclanthology.org/2023.findings-emnlp.373} {Towards making the most of chatgpt for machine translation}.
\newblock In \emph{Findings of EMNLP}.

\bibitem[{Radford et~al.(2019)Radford, Wu, Child, Luan, Amodei, Sutskever et~al.}]{radford2019language}
Alec Radford, Jeffrey Wu, Rewon Child, David Luan, Dario Amodei, Ilya Sutskever, et~al. 2019.
\newblock \href {https://insightcivic.s3.us-east-1.amazonaws.com/language-models.pdf} {Language models are unsupervised multitask learners}.
\newblock \emph{OpenAI blog}.

\bibitem[{Rei et~al.(2020)Rei, Stewart, Farinha, and Lavie}]{rei-etal-2020-comet}
Ricardo Rei, Craig Stewart, Ana~C Farinha, and Alon Lavie. 2020.
\newblock \href {https://aclanthology.org/2020.emnlp-main.213} {{COMET}: A neural framework for {MT} evaluation}.
\newblock In \emph{EMNLP}.

\bibitem[{Reimers and Gurevych(2019)}]{reimers-gurevych-2019-sentence}
Nils Reimers and Iryna Gurevych. 2019.
\newblock \href {https://aclanthology.org/D19-1410} {Sentence-{BERT}: Sentence embeddings using {S}iamese {BERT}-networks}.
\newblock In \emph{EMNLP}.

\bibitem[{Ren et~al.(2024)Ren, Zhan, Yu, Ding, and Tao}]{ren2024healthcare}
Zhiyao Ren, Yibing Zhan, Baosheng Yu, Liang Ding, and Dacheng Tao. 2024.
\newblock \href {https://arxiv.org/abs/2402.13408} {Healthcare copilot: Eliciting the power of general llms for medical consultation}.
\newblock \emph{arXiv preprint}.

\bibitem[{Robertson et~al.(2009)Robertson, Zaragoza et~al.}]{robertson2009probabilistic}
Stephen Robertson, Hugo Zaragoza, et~al. 2009.
\newblock \href {https://dl.acm.org/doi/10.1561/1500000019} {The probabilistic relevance framework: Bm25 and beyond}.
\newblock \emph{Foundations and Trends{\textregistered} in Information Retrieval}.

\bibitem[{Rubin et~al.(2022)Rubin, Herzig, and Berant}]{rubin-etal-2022-learning}
Ohad Rubin, Jonathan Herzig, and Jonathan Berant. 2022.
\newblock \href {https://aclanthology.org/2022.naacl-main.191} {Learning to retrieve prompts for in-context learning}.
\newblock In \emph{NAACL}.

\bibitem[{Schad and Engbert(2012)}]{schad2012zoom}
Daniel~J Schad and Ralf Engbert. 2012.
\newblock \href {https://www.tandfonline.com/doi/abs/10.1080/13506285.2012.670143} {The zoom lens of attention: Simulating shuffled versus normal text reading using the swift model}.
\newblock \emph{Visual Cognition}.

\bibitem[{Sia and Duh(2023)}]{sia-duh-2023-context}
Suzanna Sia and Kevin Duh. 2023.
\newblock \href {https://aclanthology.org/2023.mtsummit-research.15} {In-context learning as maintaining coherency: A study of on-the-fly machine translation using large language models}.
\newblock In \emph{MTSummit}.

\bibitem[{Socher et~al.(2013)Socher, Perelygin, Wu, Chuang, Manning, Ng, and Potts}]{socher-etal-2013-recursive}
Richard Socher, Alex Perelygin, Jean Wu, Jason Chuang, Christopher~D. Manning, Andrew Ng, and Christopher Potts. 2013.
\newblock \href {https://aclanthology.org/D13-1170} {Recursive deep models for semantic compositionality over a sentiment treebank}.
\newblock In \emph{EMNLP}.

\bibitem[{Su et~al.(2023)Su, Kasai, Wu, Shi, Wang, Xin, Zhang, Ostendorf, Zettlemoyer, Smith, and Yu}]{Selective_Annotation}
Hongjin Su, Jungo Kasai, Chen~Henry Wu, Weijia Shi, Tianlu Wang, Jiayi Xin, Rui Zhang, Mari Ostendorf, Luke Zettlemoyer, Noah~A. Smith, and Tao Yu. 2023.
\newblock \href {https://openreview.net/pdf?id=qY1hlv7gwg} {Selective annotation makes language models better few-shot learners}.
\newblock In \emph{ICLR}.

\bibitem[{Touvron et~al.(2023)Touvron, Martin, Stone, Albert, Almahairi, Babaei, Bashlykov, Batra, Bhargava, Bhosale et~al.}]{touvron2023llama}
Hugo Touvron, Louis Martin, Kevin Stone, Peter Albert, Amjad Almahairi, Yasmine Babaei, Nikolay Bashlykov, Soumya Batra, Prajjwal Bhargava, Shruti Bhosale, et~al. 2023.
\newblock \href {https://arxiv.org/pdf/2307.09288.pdf} {Llama 2: Open foundation and fine-tuned chat models}.
\newblock \emph{arXiv preprint}.

\bibitem[{Wang et~al.(2018)Wang, Singh, Michael, Hill, Levy, and Bowman}]{wang-etal-2018-glue}
Alex Wang, Amanpreet Singh, Julian Michael, Felix Hill, Omer Levy, and Samuel Bowman. 2018.
\newblock \href {https://aclanthology.org/W18-5446} {{GLUE}: A multi-task benchmark and analysis platform for natural language understanding}.
\newblock In \emph{EMNLP}.

\bibitem[{Wang and Komatsuzaki(2021)}]{gpt-j}
Ben Wang and Aran Komatsuzaki. 2021.
\newblock {GPT-J-6B: A 6 Billion Parameter Autoregressive Language Model}.
\newblock \url{https://github.com/kingoflolz/mesh-transformer-jax}.

\bibitem[{Wang et~al.(2023)Wang, Li, Dai, Chen, Zhou, Meng, Zhou, and Sun}]{wang-etal-2023-label}
Lean Wang, Lei Li, Damai Dai, Deli Chen, Hao Zhou, Fandong Meng, Jie Zhou, and Xu~Sun. 2023.
\newblock \href {https://aclanthology.org/2023.emnlp-main.609} {Label words are anchors: An information flow perspective for understanding in-context learning}.
\newblock In \emph{EMNLP}.

\bibitem[{Wang et~al.(2024{\natexlab{a}})Wang, Yang, and Wei}]{wang-etal-2024-learning}
Liang Wang, Nan Yang, and Furu Wei. 2024{\natexlab{a}}.
\newblock \href {https://aclanthology.org/2024.eacl-long.105} {Learning to retrieve in-context examples for large language models}.
\newblock In \emph{EACL}.

\bibitem[{Wang et~al.(2024{\natexlab{b}})Wang, Ding, Shen, Luo, Du, and Tao}]{wang2024oop}
Shuai Wang, Liang Ding, Li~Shen, Yong Luo, Bo~Du, and Dacheng Tao. 2024{\natexlab{b}}.
\newblock \href {https://arxiv.org/abs/2401.06628} {Oop: Object-oriented programming evaluation benchmark for large language models}.
\newblock \emph{arXiv preprint}.

\bibitem[{Ward~Bowens(2013)}]{ward2013relationship}
Saundra Ward~Bowens. 2013.
\newblock \href {https://eric.ed.gov/?id=ED553547} {The relationship between using the scrambled words reading strategy and the vocabulary of struggling readers.}
\newblock \emph{ProQuest LLC}.

\bibitem[{Wei et~al.(2022)Wei, Tay, Bommasani, Raffel, Zoph, Borgeaud, Yogatama, Bosma, Zhou, Metzler, Chi, Hashimoto, Vinyals, Liang, Dean, and Fedus}]{DBLP:journals/tmlr/WeiTBRZBYBZMCHVLDF22}
Jason Wei, Yi~Tay, Rishi Bommasani, Colin Raffel, Barret Zoph, Sebastian Borgeaud, Dani Yogatama, Maarten Bosma, Denny Zhou, Donald Metzler, Ed~H. Chi, Tatsunori Hashimoto, Oriol Vinyals, Percy Liang, Jeff Dean, and William Fedus. 2022.
\newblock \href {https://openreview.net/forum?id=yzkSU5zdwD} {Emergent abilities of large language models}.
\newblock \emph{TMLR}.

\bibitem[{Williams et~al.(2018)Williams, Nangia, and Bowman}]{williams-etal-2018-broad}
Adina Williams, Nikita Nangia, and Samuel Bowman. 2018.
\newblock \href {https://aclanthology.org/N18-1101} {A broad-coverage challenge corpus for sentence understanding through inference}.
\newblock In \emph{NAACL}.

\bibitem[{Wu et~al.(2023{\natexlab{a}})Wu, Wang, Ye, Wu, Feng, Xu, and Qiao}]{wu-etal-2023-openicl}
Zhenyu Wu, Yaoxiang Wang, Jiacheng Ye, Zhiyong Wu, Jiangtao Feng, Jingjing Xu, and Yu~Qiao. 2023{\natexlab{a}}.
\newblock \href {https://aclanthology.org/2023.acl-demo.47} {{O}pen{ICL}: An open-source framework for in-context learning}.
\newblock In \emph{ACL}.

\bibitem[{Wu et~al.(2023{\natexlab{b}})Wu, Wang, Ye, and Kong}]{wu-etal-2023-self}
Zhiyong Wu, Yaoxiang Wang, Jiacheng Ye, and Lingpeng Kong. 2023{\natexlab{b}}.
\newblock \href {https://aclanthology.org/2023.acl-long.79} {Self-adaptive in-context learning: An information compression perspective for in-context example selection and ordering}.
\newblock In \emph{ACL}.

\bibitem[{Xu et~al.(2024{\natexlab{a}})Xu, Liu, Pasupat, Kazemi et~al.}]{xu2024context}
Xin Xu, Yue Liu, Panupong Pasupat, Mehran Kazemi, et~al. 2024{\natexlab{a}}.
\newblock \href {https://arxiv.org/abs/2401.11624} {In-context learning with retrieved demonstrations for language models: A survey}.
\newblock \emph{arXiv preprint}.

\bibitem[{Xu et~al.(2024{\natexlab{b}})Xu, Peng, Ding, Tao, and Lu}]{xu-etal-2024-take-care}
Ziyang Xu, Keqin Peng, Liang Ding, Dacheng Tao, and Xiliang Lu. 2024{\natexlab{b}}.
\newblock \href {https://aclanthology.org/2024.lrec-main.1352} {Take care of your prompt bias! investigating and mitigating prompt bias in factual knowledge extraction}.
\newblock In \emph{LREC-COLING}.

\bibitem[{Zan et~al.(2022)Zan, Peng, Ding, Qiu, Liu, He, Lu, Zhang, Liu, Liu, Zhan, and Tao}]{zan2022vega}
Changtong Zan, Keqin Peng, Liang Ding, Baopu Qiu, Boan Liu, Shwai He, Qingyu Lu, Zheng Zhang, Chuang Liu, Weifeng Liu, Yibing Zhan, and Dacheng Tao. 2022.
\newblock \href {https://aclanthology.org/2022.wmt-1.37} {Vega-{MT}: The {JD} explore academy machine translation system for {WMT}22}.
\newblock In \emph{WMT}.

\bibitem[{Zhang et~al.(2024)Zhang, Ding, Zhang, and Tao}]{zhang2024intention}
Yuqi Zhang, Liang Ding, Lefei Zhang, and Dacheng Tao. 2024.
\newblock \href {https://arxiv.org/abs/2401.06561} {Intention analysis makes llms a good jailbreak defender}.
\newblock \emph{arXiv preprint}.

\bibitem[{Zhao et~al.(2021)Zhao, Wallace, Feng, Klein, and Singh}]{zhao2021calibrate}
Zihao Zhao, Eric Wallace, Shi Feng, Dan Klein, and Sameer Singh. 2021.
\newblock \href {https://arxiv.org/abs/2102.09690} {Calibrate before use: Improving few-shot performance of language models}.
\newblock In \emph{ICML}.

\bibitem[{Zhong et~al.(2023)Zhong, Ding, Liu, Du, and Tao}]{zhong2023chat}
Qihuang Zhong, Liang Ding, Juhua Liu, Bo~Du, and Dacheng Tao. 2023.
\newblock \href {https://arxiv.org/abs/2302.10198} {Can chatgpt understand too? a comparative study on chatgpt and fine-tuned bert}.
\newblock \emph{arXiv preprint}.

\bibitem[{Zhong et~al.(2024)Zhong, Wang, Xu, Liu, Ding, Du, and Tao}]{zhong2024achieving}
Qihuang Zhong, Kang Wang, Ziyang Xu, Juhua Liu, Liang Ding, Bo~Du, and Dacheng Tao. 2024.
\newblock \href {https://arxiv.org/abs/2404.14963} {Achieving >97\% on gsm8k: Deeply understanding the problems makes llms better reasoners}.
\newblock \emph{arXiv preprint}.

\end{thebibliography}

\appendix

\section{Datasets}
Natural Language Understanding (NLU) Dataset information is detailed in Table~\ref{tab:dataset}. All NLU datasets are loaded from HuggingFace Hub. For most NLU datasets, we report the results on the test set; while for the datasets MNLI and QNLI, we report the results on the validation set due to restricted access to their test sets. 

\begin{table*}[]
    \centering
    \resizebox{0.8\linewidth}{!}{
    \begin{tabular}{lcccc}
    \toprule
    \textbf{Dataset} & \textbf{Task} & \textbf{\# of Classes} & \textbf{Data Split} \\ \midrule
     \textbf{SST-2}  &  Sentiment Classification & 2 & 6920/872/1821 \\
     \textbf{SST-5}  &  Sentiment Classification & 5 & 8544/1101/2210 \\
     \textbf{CR}  &  Sentiment Classification & 2 & 3394/0/376 \\
     \textbf{Subj}  &  Subjectivity Analysis & 2 & 8000/0/2000 \\
     \textbf{AgNews} &  Topic Classification & 4 & 120000/0/7600 \\
     \textbf{MNLI}   &  Natural Language Inference & 3 & 392702/19647/19643 \\
      \textbf{QNLI}  &  Natural Language Inference & 2 & 104743/5463/5463 \\
      \bottomrule
      \end{tabular}}
    \caption{\textbf{Details of NLU datasets.}}
    \label{tab:dataset}
\end{table*}

\section{Templates}
\begin{table*}[t ]
\centering
\resizebox{0.8\linewidth}{!}{
\begin{tabular}{lll}
\toprule
\textbf{Task} & \textbf{Prompt} & \textbf{Class} \\
\hline
\multirow{2}{*}{SST-2} & Review: "<X>" Sentiment: positive & positive \\ 
& Review: "<X>" Sentiment: negative & negative \\
\midrule
\multirow{5}{*}{SST-5}
& Review: "<X>" Sentiment: terrible  & terrible \\ 
& Review: "<X>" Sentiment: bad  & bad \\ 
& Review: "<X>" Sentiment: okay  & okay \\ 
& Review: "<X>" Sentiment: good  & good \\ 
& Review: "<X>" Sentiment: great  & great \\ 
\midrule
\multirow{2}{*}{Subj}
& Input: "<X>" Type: objective  & objective \\ 
& Input: "<X>" Type: subjective  & subjective \\ 
\midrule
\multirow{2}{*}{CR} & Review: "<X>" Sentiment: positive & positive \\ 
& Review: "<X>" Sentiment: negative & negative \\
\midrule
\multirow{4}{*}{AgNews}
& "<X>" It is about world.  & World \\ 
& "<X>" It is about sports.  & Sports \\ 
& "<X>" It is about business.  & Business \\ 
& "<X>" It is about science and technology.  & Sci/Tech \\ 
\midrule
\multirow{3}{*}{MNLI}
& <C> Can we know <X>? Yes.  & Entailment \\ 
& <C> Can we know <X>? Maybe.  & Neutral \\ 
& <C> Can we know <X>? No.  & Contradiction \\ 
\midrule
\multirow{2}{*}{QNLI}
& <C> Can we know <X>? Yes.  & Entailment \\ 
& <C> Can we know <X>? No.  & Contradiction \\ 
\midrule
\end{tabular}
}
\caption{\textbf{Templates of NLU tasks.} Placeholders (e.g., <X> and <C>) will be replaced by real inputs.}
\label{tab:templates}
\end{table*}

The templates of NLU tasks used in this paper are detailed in Table~\ref{tab:templates}. For NLG tasks, we adopted templates as [src]: <X'> [tgt]: <Y'>, where [src] and [tgt] represent the source and target language names of the test language pair, respectively, and placeholders <X'> and <Y'> will be replaced by source and target sentences.

\end{document}